\pdfoutput=1
\documentclass[a4paper]{llncs}

\usepackage[cmex10]{amsmath}
\usepackage{amssymb}
\usepackage[longnamesfirst]{natbib}
\usepackage{graphicx}
\usepackage{hyperref}

\newcommand{\cov}{\mathrm{cov}}
\renewcommand{\cite}{\citep}

\begin{document}

\title{Guided Self-Organization of Input-Driven Recurrent Neural Networks}

\author{Oliver Obst\inst{1*}  \and Joschka Boedecker\inst{2*}}

\institute{%
Commonwealth Scientific and Industrial Research Organisation, Sydney, AUSTRALIA\\
\and
Machine Learning Lab, University of Freiburg, GERMANY\\ 
$^*$Both authors contributed equally to this work.
}

\maketitle

\vspace{0.7 cm}

\section{Introduction}

To understand the world around us, our brains solve a variety of tasks. One of the crucial functions of a brain is to make predictions of what will happen next, or in the near future. This ability helps us to anticipate upcoming events and plan our reactions to them in advance. To make these predictions, past information needs to be stored, transformed or used otherwise. How exactly the brain achieves this information processing is far from clear and under heavy investigation. To guide this extraordinary research effort, neuroscientists increasingly look for theoretical frameworks that could help explain the data recorded from the brain, and to make the enormous task more manageable. This is evident, for instance, through the funding of the billion-dollar "Human Brain Project", of the European Union, amongst others. Mathematical techniques from graph and information theory, control theory, dynamical and complex systems~\cite{Spo11}, statistical mechanics~\cite{RD10}, as well as machine learning and computer vision~\citep{Seu12,HB04}, have provided new insights into brain structure and possible function, and continue to generate new hypotheses for future research. 

A marked feature of brain networks is the massive amount of recurrent connections between cortical areas, especially on a local scale~\citep{DMK04}. Since information in these recurrent connections, or loops, can circulate between many neurons in a given circuit, they are ideally suited to provide a time-context for computations leading to predictions about future events. One particular mathematical model that is used to investigate the consequences of loops for computation and optimization in neuronal circuits are \emph{recurrent neural networks} (RNNs). 

In RNNs, many detailed properties of real neurons are abstracted for the sake of tractability, but important general concepts are kept. Elements of these networks are simple nodes that combine inputs from other nodes in the network, usually in a nonlinear fashion, to form their outputs. They are connected in a directed graph, which may contain cycles. In \emph{input-driven} RNNs, a constant stream of input data drives the dynamics of the network. Dedicated output units can then use this dynamics to compute desired functions, for instance for prediction or classification tasks. Since they can make use of the temporal context provided by the recurrent connections, RNNs are very well suited for time-series processing, and are in principle able to approximate any dynamical system~\cite{MJS07}.

While the recurrent connections of RNNs enable them to deal with time-dependencies in the input data, they also complicate training procedures compared to algorithms for networks without loops (e.g., Backpropagation~\cite{RHW86} or R-Prop~\cite{RB93}). Notably, training RNNs with traditional training methods suffer from problems like slow convergence and vanishing gradients. This slow convergence is due to the computational complexity of the algorithms training all of the connections in a network (such as BPTT~\cite{Wer90} or RTRL~\cite{WZ89}), as well as to bifurcations of network dynamics during training, which can render gradient information unusable~\cite{Doy92}. Also, derivatives of loss functions need to be propagated over many time steps, which leads to a vanishing error signal~\cite{BSF94,Hoc98}. 

The realization of these fundamental issues led to alternative ways of using and training RNNs, some of which can be summarized in the field of \emph{Reservoir Computing} methods~\citep[see, e.g., a recent overview by][]{LJ09}, specialized architectures like the {Long Short Term Memory} (LSTM) networks~\cite{HS97} or training by evolutionary algorithms as in the \emph{Evolino} approach~\cite{SWG+07}. The Reservoir Computing field has been an active part of RNN research over the last decade, while there was less activity in gradient-descent-like methods which appear to generate renewed interest only recently~\cite{BBP12}, partially due to the development of more efficient training techniques as in~\cite{MS11}.

Reservoir methods implement a fixed high-dimensional reservoir of neurons, using random connection weights between the hidden units, chosen small enough to guarantee dynamic stability. Input weights into this reservoir are also selected randomly, and reservoir learning procedures train only the output weights of the network to generate target outputs. A particular appeal of reservoir methods is their simplicity, and that the computation required for training is relatively low. 

\begin{figure}[tbp]
\begin{center}
\includegraphics[width=0.7\textwidth]{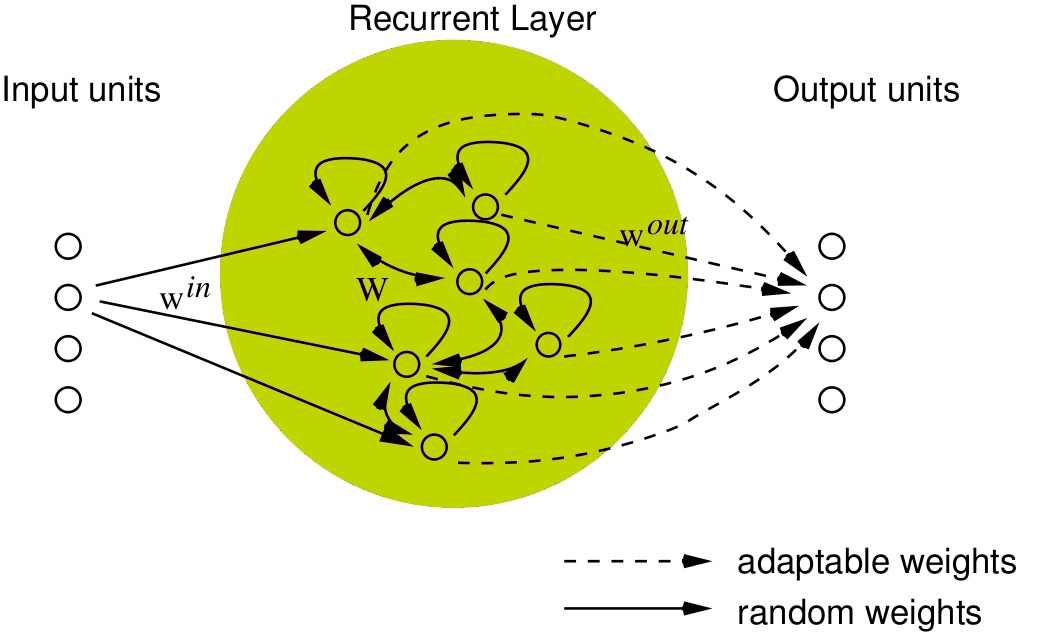}
\caption{The architecture of a typical Echo State Network (ESN), which belongs to the class of reservoir computing networks. In ESNs, the input and recurrent hidden layer (reservoir) connections are fixed randomly, and only output weights are trained. The reservoir projects the input stream nonlinearly into a high-dimensional representation, which can then be used by a linear readout layer. An important precondition for the approach to work is that the reservoir implements a fading memory, i.e. that reservoir states do not amplify, but fade out over time if no input is presented to the network.}
\label{fig:esn}
\end{center}
\end{figure}

Taking the echo state network approach as a specific example of a typical reservoir network (see Fig.~\ref{fig:esn}), it will consist of the following components:
A random input-matrix $\mathbf{W}_{in}$ combines input values $\mathbf{u}$ linearly and sends them to the units in the high-dimensional hidden layer, referred to as the \emph{reservoir}. The units in the reservoir also have recurrent connections amongst each other, collected in the matrix $\mathbf{W}_\text{res}$. These loops implement a fading memory, so information can remain in the system for some time. In this context, the metaphor of a reservoir is often used since the hidden layer can be seen as a water reservoir that gets disturbed by a drop, but slowly returns to its initial state after the ripples from the input have decayed. This reservoir state $\mathbf{x}$ is mapped at time step $t+1$ by an activation function $f()$ such as a hyperbolic tangent, in the following way:
\begin{align}
\mathbf{x}_{t+1} = f(\mathbf{W}_\text{res}*\mathbf{x}_{t} + \mathbf{W}_\text{in}*\mathbf{u}_{t+1})
\end{align}
The input and hidden layer connections, $\mathbf{W}_\text{in}$ and $\mathbf{W}_\text{res}$, are not trained in reservoir computing approaches. It is also possible to introduce feedback connections from outputs back into the reservoir~\cite{LJ09}. To approximate a specific target function, only the output weights $\mathbf{w}_\text{out}$ are trained with a simple linear regression. This drastically simplifies the training procedure compared to previous approaches, while leading to excellent performance on time-series processing tasks~\cite{JH04}. It also avoids the problems of vanishing gradient information and disruptive bifurcations in the dynamics since no error gradients have to be propagated into the fixed, random parts of the network.

This approach works very well in practice. However, results will depend on the particular random set of weights that is drawn. In fact, there is considerable variability in performance between runs of networks with equal parameter settings, but different reservoir weights drawn each time~\cite{OXP07}. Striking a balance between the two extremes of fully trained RNNs and reservoir methods, it is interesting to retain some of the simplicity and efficiency of reservoir methods, but at the same time avoid some of the variability that comes with randomly created reservoirs. Self-organized methods are of interest here, because the initial random configuration of the reservoir is in general already useful to perform the task. Each unit or connection then could, by way of local updates, contribute to an improved version of the reservoir, dependent on the data that each unit or weight processes over time. Advantages of self-organized methods are their potential for scalability, since they usually rely mainly on locally available information, making them good candidates for distributed processing.

Driving self-organization into a desired direction requires an understanding what properties a good RNN or reservoir network has. The mathematical tools to understand computation in these networks (which are instances of the larger class of input-driven dynamical systems) are still under active development~\cite{Manjunath2012}. However, different perspectives, e.g., from functional analysis, dynamical systems, information theory, or statistical learning theory already offer insights towards this goal. They can provide answers to questions such as: how well can a given RNN approximate a given class of functions? How does it implement a certain function within the collective of its distributed nodes? How much memory does an RNN provide about past inputs and where is this information stored? How does information flow through the system at different time points in time and space? How well can it separate different inputs into different network states, and how well will it generalize on data that has not been seen during training? All of these aspects contribute to the successful performance of a network on a given task (or class of tasks). Understanding how to improve them will provide possible target signals to enrich and guide the self-organized optimization process of an RNN. 

In this chapter, we review attempts that have been made towards understanding the computational properties and mechanisms of input-driven dynamical systems like RNNs, and reservoir computing networks in particular. We provide details on methods that have been developed to give quantitative answers to the questions above. Following this, we show how self-organization may be used to improve reservoirs for better performance, in some cases guided by the measures presented before. We also present a possible way to quantify task performance using an information-theoretic approach, and  finally discuss promising future directions aimed at a better understanding of how these systems perform their computations and how to best guide self-organized processes for their optimization.

\section{Assessing the Computational Power and Mechanisms of Information Processing of Reservoirs}
\label{sec:assessing}

In many cases, artificial neural networks are created with a specific goal in mind, for example to approximate a particular function or system. Training success and computational capability of the network with respect to this task are usually assessed on data that have not been used for training. Similarly to the training data, these are expected to match properties of the (yet unknown) application data well enough. A loss functional like the mean square error (MSE) or the cross-entropy is used to assess the quality of the trained system. For specific applications of the network, this is a standard approach that usually delivers meaningful results. 
When a neural network is trained for a single purpose, it is not necessary to determine its general computational power, and the loss on the validation data or during its application may be the only relevant property.

The loss on a specific class of problems does not express the general computational power of the network, though. This property becomes more interesting when a part of the system is used for more than one task: relevant cases would be dynamical reservoirs that are used for multiple applications, networks that are trained ``online'' when the task changes, or to set up or to compare generic microcircuits. One of our motivations to evaluate mechanisms of information processing is to compare self-organized approaches within reservoirs. Ideally, self-organization leads to measurable effects in the reservoir which positively affect the performance of the system. In this section, we present a number of measures for different qualities of dynamical systems that are useful in this evaluation. These measures can be roughly divided into approaches that are based on or related to information theory, approaches that relate to learning theory, and dynamical systems theory.

\subsection{Information-theory related measures}

Information theory and (Shannon) entropy have been used in a number of ways in neural network and complex systems research. One particular heuristic to measure (and eventually improve) RNN is to estimate and influence the entropy distribution of firing rates of a neuron. In individual biological neurons, for example, an approximate exponential distribution of the firing rate has been observed in visual cortical neurons~\cite{BAB+97}. Under the constraint of a fixed energy budget, i.e., a fixed mean firing rate, an exponential distribution maximizes the potentially available information: it is the maximum entropy distribution for positive random values with a given mean. \citet{Tri05} uses this idea to adapt the intrinsic excitability of a neuron with an online adaption. In this approach, the Kullback-Leibler divergence is used to measure the difference between the sample distribution of an individual neuron's output, and the exponential distribution. Target distributions different from the exponential distribution are plausible dependent on specific circumstances. For example, in reservoir networks with real-valued units, normal distributions have been used~\cite{SWV+08} to reflect the use of both negative and positive values. In both cases, the mechanism attempts to maximize the information per available energy unit locally at each neuron. 
Since energy constraints in reservoirs of artificial neural networks are typically not an issue, the maximum entropy distribution for these would in fact be the uniform. Without an energy constraint, the approach resembles the Infomax principle~\cite{Lin87}, where the average mutual information between input and output is maximized. As Bell and Sejnowski point out in their approach to maximize the mutual information for non-linear units~\cite{BS95}, for invertible continuous deterministic mappings this mutual information is maximized by maximizing the output entropy alone.
Due to the limited degrees of freedom of the approach, the desired target distribution cannot be approximated for every kind of input~\cite{Boedecker2009}. Intrinsic plasticity as well as its particular limitation can be related to Ashby's law of requisite variety~\cite{Ash56} in that by increasing variety available in the reservoir a larger variety of outputs can be successfully approximated on one hand. On the other hand, the lack of variety in the mechanism adapting the individual neurons is also responsible for the difficulty in increasing the entropy for a variety of inputs. 

The field of information dynamics~\cite{Liz07,LPZ12} provides information-theoretic measures that explicitly deal with processes or time-series. \emph{Information storage}, as one of the tools, quantifies 
how much of the stored information is actually in use at the next time step  when the next process value is computed. $A(X)$ is expressed as the mutual information between the semi-infinite past of the process $X$ and its next state $X'$, with $X^{(k)}$ denoting the last $k$ states of that process:
\begin{align}
A(X) & = \lim_{k\to\infty} A^{(k)}(X) \\
A(X,k) & = I(X^{(k)}; X') \label{eq:kapprox}
\end{align}
\emph{Information transfer}, expressed as transfer entropy~\cite{Sch00}, quantifies the influence of another process on the next state (for a formal definition, see Sect.~\ref{sec:guided} below). \citet{Boedecker2011} use these measures, to gain a better understanding of computational mechanisms inside the reservoir, and how they increase task performance for certain tasks at the phase transition between ordered and unstable dynamics. For a rote-memory task, a sharp peak can be observed in both information storage and information transfer near this phase transition, and suggests that a maximized capacity for information storage and information transfer correlates well with task performance in this dynamics regime, and this particular task. \citet{PLOW11} suggest the Fisher information matrix as a way to detect phase transitions, but this has, to our knowledge, not been applied to RNN yet.

Fisher information also plays a role in quantifying the memory stored in a dynamical reservoir: 
Information about the recent input in reservoir networks is stored in the transient dynamics, rather than in attractor states. To quantify this information storage, \citet{GHS08} use Fisher information as basis for a measure of memory traces in networks.
The measure is applicable for systems with linear activations $\mathbf{f(x)} = \mathbf{x}$, subject to Gaussian noise $\mathbf{z}$, and input $v(t)$:
\begin{align}
 \mathbf{x}(t) & = \mathbf{f}(\mathbf{W}^\text{in} v(t) + \mathbf{Wx}(t-1) +  \mathbf{z}(t)).
\end{align}
The Fisher Memory Matrix (FMM) between the present state of the system $\mathbf{x}$ and the past signal is defined as\begin{align}
J_{k,l}(\mathbf{v}) & = 
\left\langle - \frac{\delta^2}{\delta_\mathbf{v_k} \delta_\mathbf{s_l}} \log P(\mathbf{x}(t)|\mathbf{v}) 
\right\rangle_{P(\mathbf{x}(t)|\mathbf{v})}.
\end{align}
Diagonal elements $J(k) \equiv \mathbf{J}_{k,k}$ are the Fisher information that the system keeps in $\mathbf{x}(t)$ about  a pulse 
at $k$ steps back in the past, i.e., the decay of the memory trace of a past input. $J(k)$ is called the Fisher memory curve (FMC). 
\citet{TR13} investigate the relation between $J(k)$ and the short term memory capacity MC~\cite{Jae01b} (details on MC in the following subsection), and show that the two are closely related in linear systems. For these, $J(k)$ is independent of the actual input used. In the general, nonlinear case that is interesting for us, however, the FMC depends on the input signal, as the memory capacity MC does, and is hard to analyze. 
 
A measure for Active Information Storage in input-driven systems has been proposed to quantify storage capabilities of a nonlinear system independent of particular inputs~\cite{OBSA13b}. The measure is an Active Information Storage~\cite{LPZ12} where the current input $u_{n+1}$ is conditioned out:
\begin{align}
A^U_X(k) & = \left\langle  a_X^U(n+1,k) \right\rangle_n \,\text{, with} \\
a_X^U(n+1,k) & = \log \frac{p(x_n^{(k)}, x_{n+1}|u_{n+1})} {p(x_n^{(k)}) \, p(x_{n+1}|u_{n+1})} \\
 & = \log \frac{p(x_{n+1}|x_n^{(k)}, u_{n+1})} {p(x_{n+1}|u_{n+1})}.
\end{align}

The idea for this measure is to remove influences of structure in input data, and to only characterize the system itself, rather than a combination of system and input data. In theory, this influence would be removed by having the history sizes in computing the information storage converge to infinity. Large history sizes, however, require large amounts of data to estimate the involved joint probabilities, and this data, and the time required for the estimation is often not available. Active information Storage for input-driven systems assesses one aspect of the computational capabilities of a dynamical system, others, like the information transfer, would need to be defined for input-driven systems in a similar way. 
 
\subsection{Measures related to learning theory}

\citet{Legenstein2007} propose two measures to quantify the computational capabilities of reservoirs in the context of liquid state machines, one of the two main flavors of reservoir computing networks: the linear separation property und the generalization capability. The linear separation property quantifies the ability of a computational system to map different input streams to significantly different internal states. This is useful because only then will the system be able to (linearly) map internal states to different outputs. The measure is based on the rank of an $n \times m$ matrix $M$ whose columns are state vectors $\mathbf{x}_{u_{i}}(t_{0})$ of circuit $C$ after having been driven by input stream $u_{i}$ up to a fixed time $t_{0}$. These state vectors are collected for $m$ different input streams, i.e., $u_{1},\ldots, u_{m}$. If the rank of $M$	 is $m$, then $C$, together with a linear readout, is able to implement any assignment of output units $y_{i} \in \mathbb{R}$ at time $t_{0}$ for inputs $u_{i}$.	

For the generalization ability, they propose to approximate the VC-dimension of class ${\cal H}_{C}$ of the reservoir, which includes all maps from a set $S_\text{univ}$ of inputs $u$ into $\{0,1\}$ which can be implemented by a reservoir $C$. They present a theorem (and corresponding proof sketch) stating that under the assumption that $S_\text{univ}$ is finite and contains $s$ inputs, the rank $r$ of a $n \times s$ matrix whose columns are the state vectors $\mathbf{x}_{u}(t_{0})$ for all inputs $u$ in $S_\text{univ}$ approximates the VC-dimension(${\cal H}_{C}$), specifically $r \le \text{VC-dimension}({\cal H}_{C}) \le r+1$.

According to~\cite{Legenstein2007},  a simple difference of both (normalized) measures leads to good predictions about which reservoirs perform well on a range of tasks. 

The loss or the success on a set of test functions is another possibility to characterize the systems from a learning point of view. One such measure is the short term memory capacity MC~\cite{Jae01b} that we briefly mentioned above. To compute the MC, a network is trained to generate delayed versions $v(t-k)$ of a single channel input $v(k)$. The measure then is the sum of the precisions for all possible delays, expressed as a correlation coefficient:
\begin{align}
\mathrm{MC} & =  \sum_{k=1}^\infty \mathrm{MC}_k \\
\mathrm{MC}_k & = \max_{\mathbf{w}_k^\text{out}} \frac{\cov^2(v(t-k), y_k(t))} {\sigma^2(v(t)) \, \sigma^2(y_k(t))}, \,\text{with}\\
y_k(t) & = \mathbf{w}_k^\text{out} \, \binom {v(t)} {\mathbf{x}(t)}  \text{, and } \, 
\mathbf{x}(t) = \mathbf{f}(\mathbf{W}^\text{in} v(t) + \mathbf{Wx}(t-1)). \nonumber
\end{align}

The symbols $\cov$ and $\sigma^2$ denote covariance and variance, respectively. Each coefficient takes values between 0 and 1, and expresses how much of the variance in one signal is explainable by the other. As shown in \cite{Jae01b}, for i.i.d. input and linear output units, the MC of $N$-unit RNN is bounded by $N$. The measure is related to the Fisher memory matrix approach above. 

Another approach in this area is the \emph{information processing capacity} of a dynamical system~\cite{Dambre2012}. It is a measure based on the mean square error MSE in reconstructing a set of functions $z(t)$. The idea is to distinguish from approaches that view dynamical systems merely providing some form of memory for a -- possibly nonlinear -- readout. In \cite{Dambre2012}, systems are regarded as both providing memory and performing nonlinear computation. The readouts then only combine states of the system linearly, attempting to minimize the MSE for a function $z(t)$, so that all essential aspects of computation have to be covered by the dynamical system. The capacity of the system for approximating the desired output is computed using the (normalized) MSE of the optimal linear readout,  
\begin{align}
C_T[X,z] & = 1 -  \frac{\min_W \text{MSE}_T[\hat{z}]}{\langle z^2 \rangle_T}
\end{align}

This computed capacity is dependent on the input. In order to avoid an influence of structure in the input 
on the results, i.i.d.\ input is required for the purpose of measuring the capacity. To measure information processing capacity, several functions $z$ have to be evaluated. The idea is that if $z$ and $z'$ are orthogonal, $C_T[X,z]$ and $C_T[X,z']$ measure independent properties of the system. The total capacity, on the other hand, is limited by the number of variables $x_i$, so that a finite number of output functions is sufficient. A possible choice of output functions are Legrende polynomials, which are orthogonal over $(-1,1)$. 

The proposed approach has been used to compare different implementations of dynamical systems, like reaction-diffusion systems and reservoirs. An interesting idea that is also mentioned in~\cite{Dambre2012} would be to extend the approach so that the underlying system adapts to provide specific mappings. One possibility might be to adjust the number of internal units in an online-learning setting, e.g., when the task changes. The requirement for i.i.d.\ input is a limitation of the current approach, though it appears that even in the non-i.i.d.\ input case useful information about the system can be gathered. It might also be interesting to compare how the approach relates to the information-dynamics framework~\citep{Liz07,LPZ12} to quantify computation in non-linear systems.

\subsection{Measures related to dynamical systems theory}

To gain understanding of the internal operations that enable high-dimensional RNNs solving a given task,
a recent effort by~\citet{Sussillo2013} draws on tools from dynamical systems theory.
Using numerical optimization techniques, the authors identify fixed points and points of only gradual change (also called \emph{slow points}) in the dynamics of the networks. Linearization around these points then reveals computational mechanisms like fixed points that implement memories, saddle points (fixed points with one unstable dimension) that enable switching between memory states, and approximate plane attractors that organize the storage of two consecutive input values to be memorized. For the tasks that were looked at in this work, the computational mechanisms could be inferred from the linearized dynamics around the set of fixed and slow points, and task performance of the trained networks was well explained. 

In~\cite{Williams2010}, the authors argue for a complementary role of dynamical analysis, which involves, e.g., looking at attractors and switching between attractor states, and also an information-theoretic analysis when trying to understand computation in dynamical systems (including input-driven ones -- even though the input might simply be considered as part of the environment and is assumed to be distributed uniformly). They evolve agents that are controlled by small continuous-time recurrent neural networks (CTRNNs) and evaluate their behavior in a relational categorization task. This involves keeping a memory about different objects the agent can sense, and reacting with avoid or catch behaviors based on the relation of both objects. Dynamical analysis shows that the state of a specific neuron in the CTRNN is 
correlated with the size of the first object, and switching on or off a different neuron determines whether the agent catches or avoids the second object. Both features are found to be connected through a saddle-node bifurcation in the CTRNN dynamics whose timing and location depends on properties of the second object. The desire to understand the flow of information through the brain-body-environment system 
between these events leads the authors to information-theoretic measures unrolled over time (similar to the motivation and approach in~\cite{Liz07}). By considering the temporal evolution of measures like conditional mutual information, they are able to measure \emph{information gain} or \emph{information 
loss} of a state variable at specific time points. Similarly, they can quantify the \emph{specific information} 
that a state variable carries about a particular stimulus at each time step. The behavior of the agent can then be explained by a sudden gain and then loss of information about object sizes in the first neuron, and then a rapid gain of information about relative size of the objects. In summary, the authors state that the two different ways to look at the computational mechanisms of the RNN differ, but provide coherent and even complementary information on how the agent solves the task that would be difficult to get with either approach alone. 

Another approach from dynamical systems theory to understand and predict computational capabilities in RNNs builds on the concept of Lyapunov exponents. Although these concepts are only defined for autonomous dynamical systems, an analogous idea is to introduce a small perturbation into the state of one of two identical networks but not the other, and observe the time evolution of the state difference while the networks are driven with identical input. In case the perturbation fades out, the network is assumed to be in the stable phase of the dynamics. If it amplifies, the network is in the unstable, and possibly, the chaotic dynamics regime. If it approximately persists, the network is arguably at the phase-transition between stable and unstable regimes. Example applications of this approach can be found in~\cite{BN04,Boedecker2011}. In~\cite{BN04}, it was observed that the performance of binary threshold unit RNNs is maximized at this phase-transition for a task that requires memory and nonlinear processing to be solved successfully. This result was later replicated for analogue Echo State Networks in~\cite{Boedecker2011} for a rote-memory task; however, it was also found that some tasks do not benefit from reservoirs at the phase-transition, as observed before in the complex systems literature~\cite[e.g.,][]{MHC93}.

\section{Improving Reservoir Information Processing Capabilities Through Self-Organized Adaptation}
\label{sec:improving}

A pragmatic way to evaluate the quality of a reservoir is to train the output, and evaluate it on a training or validation set~\cite{Luk12}. In most circumstances, training is fast so that a number of hyper-parameter settings can be tested. \citet{Luk12} proposes a number of invaluable recipes to reservoir production. The recipes are very helpful for creating a good enough reservoir before output weights are trained. They show up promising directions for exploration, but are intended to be used as a guide rather than hard and fast rules, as some of them are mutually exclusive. The approaches selected for this section are intended to improve the reservoir itself in a self-organized way after it was created or selected. Possibly, this might happen simultaneously in combination with online learning of output weights, or, alternatively, as a self-organized pre-training approach followed by the standard offline output weight training. 

%

\subsection{SORN: self-organized optimization based on 3 local plasticity mechanisms}
One approach that has demonstrated how self-organization can be leveraged to optimize a reservoir network can be found in~\cite{Lazar2009}. SORN is a self-organizing recurrent network architecture using discrete-time binary units. The three plasticity mechanisms are: a variant of spike-time dependent plasticity (STDP), adjusting certain weights in the reservoir, a synaptic normalization rule (SN) responsible to keep the sum of afferent weights of a neuron constant, and IP learning to adapt the unit firing threshold. The network state evolves using the following update functions:
\begin{align}
R_i(t+1) &= \sum_{j=1}^{N^E} W_{ij}^{EE}(t) x_j(t) - \sum_{k=1}^{N^I} W_{ik}^{EI} y_k(t) - T_i^E(t)\\ \label{eq:recurrent}
x_i(t+1) &= \Theta(R_i(t+1) + v_i^U(t))  \\
y_i(t+1) & = \Theta(\sum_{j=1}^{N^E} W_{ij}^{IE}(t) x_j(t) - T_j^I)
\end{align}

$T^E$ and $T^I$ are threshold values, drawn randomly from positive intervals for excitatory units and inhibitory units, respectively. $\Theta$ is the heaviside step function, and $v_i^U(t)$ the network input drive. Matrices $W^{IE}$ and $W^{EI}$ are fully connected, and represent connections between inhibitory and excitatory units, and vice versa. $W^{EE}$ holds connections between excitatory units. These are random, sparse, and without self-recurrence. Inhibitory units are not directly connected to each other. All weights are drawn from the interval $[0,1]$, and the three matrices $W^{IE}$, $W^{EI}$, and $W^{EE}$ are normalized, i.e., $\sum_j W_{ij} = 1$. The network state at time $t$ is given by the two binary vectors $x(t) \in \{0,1\}^{N^E}$, and $y(t) \in \{0,1\}^{N^I}$, representing activity of the $N^E$ excitatory and the $N^I$ inhibitory units, respectively. 

STDP and synaptic scaling update connections of excitatory units of the reservoir, while IP changes their thresholds. Inhibitory neurons and their connections remain unchanged. In the SORN the STDP for some small learning constant $\eta_{stdp}$ is formalized as:
\begin{align}
\Delta W_{ij}^{EE}(t) & = \eta_{stdp}(x_i(t)x_j(t-1) - x_i(t-1)x_j(t)).
\end{align}
Synaptic scaling normalizes the values to sum up to one:
\begin{align}
\Delta W_{ij}^{EE}(t) & = W_{ij}^{EE} (t) / \sum_j W_{ij}^{EE}(t).
\end{align}
IP learning is responsible for spreading activations more evenly, using a learning rate $\eta_{ip}$, and a target firing rate of $H_{IP}$:
\begin{align}
T_i^E(t+1) &= T_i^E(t) + \eta_{ip}(x_i(t)-H_{IP}
\end{align}

\begin{figure}[tbp]
\begin{center}
\includegraphics[width=\textwidth]{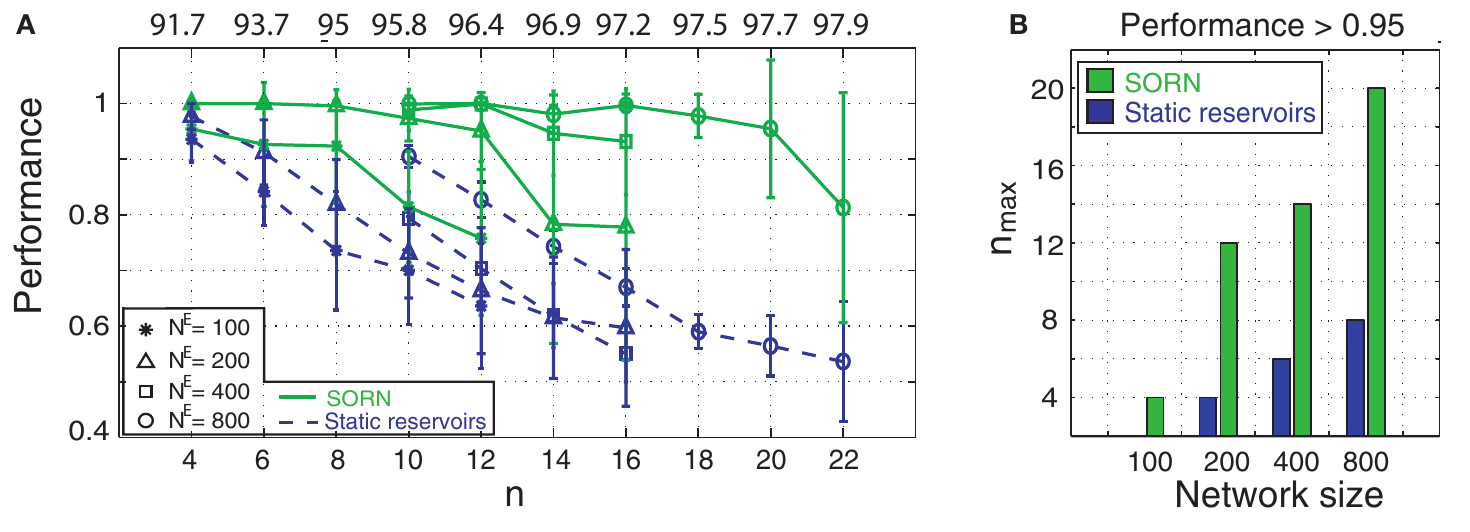}
\caption{(left) Normalized performance versus task difficulty as indicated by $n$, the number of repeated characters of a word which the network should predict. Different network sizes were tested. The numbers on top indicate the maximum possible performance -- which is limited by the inherent randomness of the first character of a word within the sequence. Standard deviation among trials is indicated by the error bars. (right) Highest value of $n$ for which a network achieved more than $95\%$ of maximum performance as function of network size. The plastic SORN networks are able to deal with significantly harder tasks than the static reservoirs at this performance level. Graphs reproduced from~\cite{Lazar2009}.} 
\label{fig:sornperformance}
\end{center}
\end{figure}

\citet{Lazar2009} show that the SORN outperforms static reservoir networks using a letter prediction task. The network has to predict the next letter in a sequence of two different artificial words of length $n+2$. These words are made up of three different characters, with the second character repeated $n$ times. The first character of a word is random and the network cannot do better than randomly guessing which one will come up. If the reservoir is able to efficiently separate the repeated character in the middle part of the word, though, the network can learn to count these characters and predict the rest of the sequence correctly. Figure~\ref{fig:sornperformance} compares the normalized performance of SORNs and static reservoir networks of different sizes on instances of the task with increasing $n$ (increased difficulty). SORNs are able to outperform static reservoirs clearly on this task. A PCA analysis in~\cite{Lazar2009} reveals that the SORN indeed shows a much better separation property and maps repeated inputs to distinct network states, while the states of static reservoirs are much more clustered together and thus harder to distinguish by the linear readout. 

\begin{figure}[tbp]
\begin{center}
\begin{minipage}{0.49\textwidth}
\includegraphics[width=\textwidth]{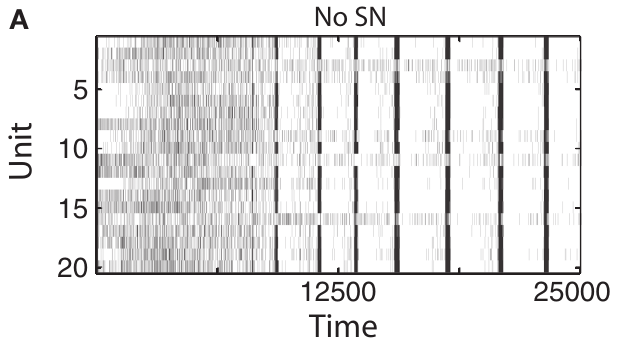}
\end{minipage}
\hfill
\begin{minipage}{0.49\textwidth}
\includegraphics[width=\textwidth]{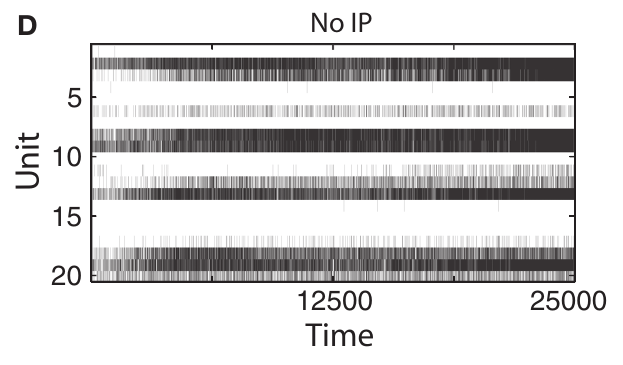}
\end{minipage}
\caption{Activity snapshots for 50 randomly selected reservoir neurons. (left) A reservoir without synaptic scaling develops highly synchronized, seizure-like firing patterns. (right) Without the IP mechanism, neuron activity is unevenly distributed with some neurons firing almost constantly while others are nearly silent.  Graphs reproduced from~\cite{Lazar2009}.} 
\label{fig:sicksorn}
\end{center}
\end{figure}

The combination of the three mechanisms appears to be a key to successful self-organization in an RNN. Figure~\ref{fig:sicksorn} illustrates that the dynamics of SORN reservoir become sub-optimal if only two of the three plasticity mechanisms are active. Without synaptic normalization, the network units become highly synchronized. This severely restricts the representational power of the reservoirs. If IP learning is switched off, the activity of neurons in the network becomes unbalanced. Some neurons fall nearly silent while others are active almost all the time. This is in contrast to the case with IP where activity is more evenly distributed, enabling a richer representation of information in the reservoir.

Though some of the self-organizing mechanisms like STDP are biologically plausible, there are not too many examples of successful applications for training RNNs, or, as \citet{Lazar2009} states, ``Understanding and controlling the ensuing self-organization of network structure and dynamics as a function of the network's inputs is a formidable challenge''. For time-series prediction and system identification tasks, an extension of the approach to analog units would be required. Also, an investigation of the information dynamics during and after adaptation may provide insights, for example into the relation between reservoir configuration and information transfer.

\subsection{Hierarchical Self-Organizing Reservoirs} 
A different approach based on self-organized optimization of reservoirs is presented in~\cite{Luk12b}. The author compares classical ESNs and recurrent RBF-unit based reservoir networks (called Self-Organizing Reservoirs, SORs) which resemble Recurrent Self-Organizing Maps (RSOMs)~\cite{Voe02}. The input and reservoir weights of the SOR are adapted by learning rules traditionally used for Self-Organizing Maps (SOMs)~\cite{Koh01} and NeuralGas networks~\cite{MS91}. 

The update equations for the SOR are:
\begin{align}
\tilde{x}_{i}(n) & = \exp(-\alpha \Vert \mathbf{v}^{in}_{i} - \mathbf{u}(n) \Vert^{2} - \beta \Vert \mathbf{v}_{i} - \mathbf{x}(n-1) \Vert^{2}), \quad i = 1,\ldots,N_{x},\\
\mathbf{x}(n) & = (1 - \gamma)\mathbf{x}(n-1) + \gamma\mathbf{\tilde{x}}(n).
\end{align}

Here, the internal reservoir neuron states at time $n$ are collected in vector $\mathbf{x} \in \mathbb{R}^{N_{x}}$ and their update in vector $\mathbf{\tilde{x}} \in \mathbb{R}^{N_{x}}$. The factor $\gamma \in (0,1]$ is the leak-rate. The vector $\mathbf{u} \in \mathbb{R}^{N_{u}}$ contains the input-signal, while matrices $\mathbf{V}^{in}$ and $\mathbf{V}$ are the input and recurrent weight matrix, respectively, whose $i$th column vectors are denoted by $\mathbf{v}^{in}_{i} \in \mathbb{R}^{N_{x}}$ and $\mathbf{v}_{i} \in \mathbb{R}^{N_{x}}$. Parameters $\alpha$ and $\beta$ scale the input and recurrent distances, and $\Vert \cdot \Vert$ denotes the Euclidean norm.

The unsupervised training of the SOR updates the input and recurrent weights as:
\begin{align}
\mathbf{v}^{all}_{i}(n+1) & = \mathbf{v}^{all}_{i}(n) + \eta (n)h(i,n)([\mathbf{u}(n) ; \mathbf{x}(n)] - \mathbf{v}^{all}_{i}(n)), 
\end{align}

where $\mathbf{v}^{all}_{i}(n) \equiv [\mathbf{v}^{in} ; \mathbf{v}]$ and $\eta (n)$ is a time-dependent learning rate. The learning-gradient distribution function $h$ is defined either as:
\begin{align}
h(i,n) =  \exp(-d_{h}(i,\text{bmu}(n))^{2} / b_{h}(n)^{2}),
\end{align}
 
 where $d_{h}(i,j)$ is the distance between reservoir units $i$ and $j$ on a specific topology, $\text{bmu}(n) = \arg \max_{i}(x_{i}(n))$ is a function returning the index of a \emph{best matching unit} (BMU), and $b_{h}(n)$ is the time-dependent of the learning gradient distribution. With this definition of $h$, the learning proceeds according to the SOM algorithm. To implement NeuralGas-like learning, it suffices to change this definition to:
 \begin{align}
h_{ng}(i,n) =  \exp(-d_{ng}(i,n) / b_{h}(n)),
\end{align}

where $d_{ng}(i,n)$ denotes the index of node $i$ in the descending ordering of activities $x_{i}(n)$ (see~\cite{Luk12b} for additional details). Both algorithms were found to be similarly effective to improve the pattern separation capability of reservoirs compared to standard ESNs when tested on detection of certain signal components on a synthetic temporal pattern benchmark, and on classification of handwritten digits from a stream of these characters. Further improvements are reported if these SORs are stacked on top of each other in a hierarchy, trained in a layer-by-layer fashion. However, results only improve if enough time is given for the self-organization process to find suitable representations. If layers are stacked with very little training time for each of them, performance actually worsens. 

\subsection{Guided Self-Organization of Reservoir Information Transfer}
\label{sec:guided}
In~\cite{Obst2010}, the information transfer between input data and desired outputs is used to guide the adaptation of the self-recurrence in the hidden layer of a reservoir computing network. The idea behind this step is to change the memory within the system with respect to the inherent memory in input and output data (see Section~\ref{sec:taskcomplexity} below for a a discussion which develops these ideas further). 

The network dynamics is updated as:
\begin{align}
\mathbf{x}(k + 1) & = diag(\mathbf{a})\mathbf{Wy}(k) + (\mathbf{I} - diag(\mathbf{a}))\mathbf{y}(k) + \mathbf{w}^{in}\mathbf{u}(k)\\
\mathbf{y}(k + 1) & = \mathbf{f}(\mathbf{x}(k + 1)) ,	
\end{align}
where $x_{i}, i = 1, \ldots, N$ are the unit activations, $\mathbf{W}$ is the $N \times N$ reservoir weight matrix, $\mathbf{w}^{in}$ the input weight vector, $\mathbf{a} = [a_{1}, \ldots, a_{N} ]^{T}$ a vector of local decay factors, $\mathbf{I}$ is the identity matrix, and $k$ denotes the discrete time step. As a nonlinearity, $f(x) = \tanh(x)$ is used.
The $a_{i}$ represent the coupling of a unit's previous state with the current state, and are computed as:
\[
a_{i} = \frac{2}{1+m_{i}},
\]
where $m_{i}$ represents the memory length of unit $i$ ($m_{i} \in \{1, 2, 3, \ldots \}$), initialized to $m_{i} = 1$. Increasing individual $m_{i}$ through adaptation increases the influence of a unit's past states on its current state. The information transfer is quantified as a conditional mutual information or transfer entropy~\cite{Sch00}:
\begin{align}
	T_{X\to Y} & = \lim_{k,l\to\infty} T_{X\to Y}^{(k,l)}, \text{with}\\
	T_{X\to Y}^{(k,l)} & = I(X^{(l)};Y'|Y^{(k)}).
\end{align}
Parameters $k$ and $l$ are history sizes, which lead to finite-sized approximations of the transfer entropy for finite values.

In a first step, the required history size $l$ is determined which maximizes the information transfer $T_{u \to v}$ from input $u$ to output $v$. This value will  increase for successively larger history sizes, but the increases are likely to level off for large values of $l$. Therefore, $l$ is determined as the smallest value which is still able to increase $T_{u \to v}$ by more than a threshold $\epsilon$:
\begin{align}
T_{\mathbf{u} \to \mathbf{v}}(1, \hat{l} + 1) & \leq T_{\mathbf{u} \to \mathbf{v}}(1, \hat{l}) + \epsilon \quad \text{and}\\
T_{\mathbf{u} \to \mathbf{v}}(1,l) & > T_{\mathbf{u} \to \mathbf{v}}(1, l - 1) + \epsilon \quad \text{for all} \; l < \hat{l}.	
\end{align}

In a second step, the local couplings of the reservoir units are adapted so that the transfer entropy from the input of each unit to its respective output is optimized \emph{for the particular input history length $\hat{l}$}, as determined in step one. Over each epoch $\theta$ of length $\ell$, we compute the transfer entropy from activations $x_{i}^{(\ell)}$ to output $y_{i}^{(\ell)}$ for each unit $i$:
\[
te_{i}^{\theta} =T_{x_{i}^{(\ell)} \to y_{i}^{(\ell)}}(1,\hat{l}).
\]
If the information transfer during the current epoch $\theta$ exceeds the information transfer during the past epoch by a threshold (i.e., $te_{i}^{\theta} > te_{i}^{\theta - 1} + \epsilon$), the local memory length $m_{i}$ is increased by one. In case $te_{i}^{\theta} < te_{i}^{\theta - 1} - \epsilon$, the local memory length is decreased by one, down to a minimum of $1$. The decay factors $a_{i}$ are fixed once they stabilize, which ends the pre-training phase.

\begin{figure}[tbp]
\begin{center}
\includegraphics[width=\textwidth]{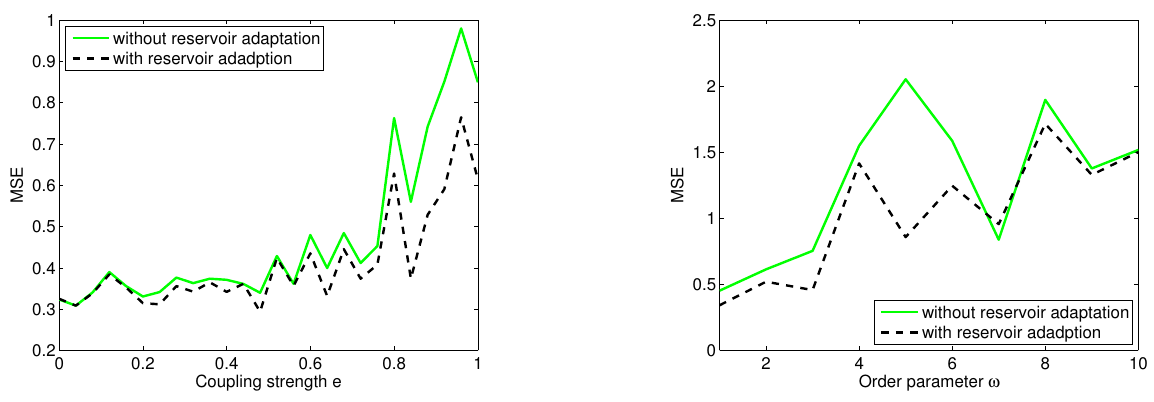}
\caption{(left) Mean squared errors of the prediction over the test data for different coupling strengths and fixed $\omega = 0$. (right) Mean squared error for different $\omega$ using a fixed coupling of $e = 0.75$. }
\label{fig:TEresimprov}
\end{center}
\end{figure}

In~\cite{Obst2010}, the method is tested on a one-step ahead prediction of unidirectionally coupled maps and of the Mackey-Glass time series benchmark. Showing results for the former task as an example, Figure \ref{fig:TEresimprov} (left) displays the mean square errors of the prediction over the test data for different coupling strengths $e$ and fixed order parameter $\omega$ for both echo state learning with and without adaptation of information transfer in the reservoir (averages over 50 trials). For each individual trial the same reservoir and time series have been used once with and without adaptation. The prediction using the reservoir adaptation is better over almost the entire range of $e$, with the improvement becoming more significant as the influence of the input time series becomes larger. Figure \ref{fig:TEresimprov} (right) is a plot of the mean square error for different $\omega$ using a fixed coupling $e$. In all but one cases the reservoir adaptation improves results.

\section{Quantifying task complexity}
\label{sec:taskcomplexity}

Most currently existing measures capture some of the generic computational properties of recurrent neural networks (as an important class of input-driven system), such as memory capacity or entropy at the neuron-level, but do not take task complexity into account. 
Optimization of the network properties based on these generic measures therefore will only do a ``blind'' adjustment of parameters while no optimality  guarantee for the task at hand can be given. 
More positively put, the philosophy behind these measures is that a maximization of some of them leads to reservoirs that are capable to solve a variety of tasks. The self-organizing mechanisms in Sect.~\ref{sec:improving} are one way to achieve this maximization. 
In situations where no teaching signal is given, e.g., in clustering tasks, one can do no better than that; however, if the desired output signal is available, it can be used to quantify the task complexity as a relationship between inherent difficulty of predicting the output based on its own history, and to what extent the input data can contribute to improve these predictions. 
This would inform us how achievable a task is, and may also be used to trade off complexity of the system against the expected quality of the solution.


It is possible to use some of the tools that we introduced above, and take an information-theoretic approach to tackle this problem. Essentially we are interested in quantifying how difficult it is for a system to produce its next output. The systems we are interested in take a time series $X$ as an input, and have the goal to generate output $Y$, another time series. To produce the next state $y_{t+1}$, both the output's past ($y_1 ... y_t$) as well as the input up to the current step ($y_1, ..., y_{t+1})$ can be considered. 

The Active Information Storage $A_Y$~\cite{LPZ12} can be used to capture the influence of previous outputs in producing the next output: how much information is contained in the past of $Y$ that can be used to compute its next state? This is expressed as the average mutual information between past $Y^{(k)}$ of the output and the next state $Y'$:
\begin{align}
	A_Y & = \lim_{k\to\infty} A_Y(k), \text{with}\\
	A_Y(k) & = I(Y^{(k)};Y').
\end{align}
We use $A_Y(k)$ to represent finite-$k$ estimates. Now, $A_Y$ and $A_Y(k)$ allow for two kind of measurements: (a) higher values for $A_Y$ indicate better predictability of $Y$ from its own past, i.e., $A_Y$ is one component of the overall task difficulty.  (b) With increasing values of $k_i = 1,2,...,n$, estimates $A_Y(k_i)$ indicate the amount of memory that is in use. As the information that can be used to predict the next state increases with larger values of $k_i$, $A_Y(k_i)$ will monotonically increase with $k_i$, and asymptotically converge to some maximum. Finding $k^*$ so that $A_Y(k^*) \geq  A_Y(k) +\epsilon$, for some small $\epsilon > 0$ and $k\to\infty$, thus gives us a useful quantity for the amount of memory required, and at the same time $A_Y(k^*) \approx A_Y$  quantifies the difficulty in predicting $Y'$ from its own past.

The other component that plays a role in the task is the input $X$. Its contribution to producing the next output $Y'$ of the system, too, can be quantified, using the transfer entropy introduced above.
The transfer entropy indicates how much information the input $X$ contributes to the next state $Y'$, given that the past of $Y$ is known. Increasing the input history size $l$ increases the information available in computing $Y'$, for fixed output history size $k^*$.  Large $T_{X\to Y}$ suggest that the input $X$ helps in computing the next output, i.e., the task for the system is less difficult than for smaller transfer entropies.  Finding an input history size $l^*$ so that $T_{X\to Y}^{(k^*,l^*)} \approx \lim_{l\to\infty} T_{X\to Y}^{(k^*,l)}$ gives us another useful quantity for the amount of memory required. 

Unfortunately, using these quantities to compare tasks or to design systems is not entirely straightforward, for a number of reasons. For continuous-valued time series $X$ and $Y$, estimating mutual informations is cumbersome, and requires larger amounts of data in particular for larger history sizes $l$ and $k$. To compare task difficulties, it would also be helpful to normalize both quantities, e.g., for the Active Information Storage by the joint entropy $H(Y;Y')$, to values between 0 and 1. The true output history $Y$ may also be simply not available to the system, dependent on how it operates. For example, in batch mode, the only information that is available is the input $X$ and the \emph{estimated} output $\hat{Y}$. The true history of $Y$ is usually only accessible if the system operates online.
Finally, the two components $A_Y$ and $T_{X\to Y}$ cannot be simply added to specify the overall task difficulty since input and output may redundantly share some information.

We will reserve a detailed investigation of applying both measures to a later publication. As a concept to explain contributions of input and output history, they can be an indicator for how complex the information processing system needs to be. It will also be interesting to see how other measures relate to them, and to show which aspect of the computation they measure. As an example, 
the memory capacity MC, as a sum of correlation coefficients can be seen as a linear measure of the potential information transfer between input $X$ and the desired output $Y$. Mutual information expresses a nonlinear relationship between two variables, and so does the transfer entropy, a conditional mutual information between $X$ and $Y$. 
In contrast to MC, the $T_{X\to Y}$ measures actual information transfer between input and desired output, i.e., $T_{X\to Y}$ is a purely a property of the task. MC is a property of the RNN, but as it is using task specific input, it combines the properties of the RNN with properties of the input. The two quantities could be used to adjust the architecture of a neural network for better performance on a specific task. 

As another example, measuring the individual distributions of unit activations in the reservoir and their divergence from a maximum entropy distribution capture properties of the input combined with properties of the network. On the other hand, Active Information Storage for input-driven systems, applied to reservoirs or individual reservoir units, expresses the amount of information in the system that is in use to predict the next state, and is meant to measure capabilities of the system only. 

Related work on complexity measures includes Grassberger's forecast complexity~\cite{Gra86,Gra12}, which considers the difficulty of making an optimal prediction of a sequence created by a stochastic process. A sequence can be compressed up to its entropy rate, and the forecast complexity is the computational complexity of the algorithm responsible for the decompression. The probability of the next symbol is needed for this decompression. Related ideas can be found in Minimum Description Length (MDL) approaches \cite{Ris78} and Kolmogorov complexity~\cite{Kol65} as measures for complex objects. 
Also introduced by \citet{Gra86,Gra12} is the Effective Measure Complexity EMC, the relative memory required to calculate the probability distribution of the next symbol of a sequence. The EMC is a lower bound on the forecast complexity. 
Both forecast complexity and EMC look at sequences, e.g., the output of an autonomous system without regard to its input, whereas we are interested in systems that produce an output based on some input. 
More complexity measures  can be found in a special issue on ``Measures of Complexity from Theory to Applications'', with \cite{CM11} as an introductory article.

\section{Conclusion}



We presented methods to assess different computational properties of input-driven RNNs, and reservoir computing networks in particular, in the first part of the paper. These methods were drawn from information-theory, statistical learning theory, and dynamical systems theory, and provided different perspectives on important aspects of information processing in these systems. They help to quantify properties like the memory capacity a certain network provides, the flow of information through the system and its modification over time, the ability to separate similar inputs and generalize to new, unseen data, and others. In addition to their usefulness in their own right when trying to understand how RNNs implement the functions they are trained for, they also have the potential to be used as target signals to guide self-organized optimization procedures aimed at improving the quality of reservoirs for a specific task over random initialization.

In the second part of the paper, we presented some recent efforts at implementing self-organized optimization for reservoir computing networks. One approach combined different plasticity mechanisms to improve coding quality and separation ability of the network, while a different approach was using methods similar to recursive self-organizing maps with SOM and NeuralGas-like learning rules. The final approach we presented proceeds in two phases: determining a learning goal in terms of information transfer between input and desired output, and using this quantity to guide local adjustments to the self-recurrence of each reservoir unit. All of these methods showed the potential of self-organized methods to improve network performance over standard, random reservoirs while avoiding problems associated with back propagation of error-gradients throughout the whole networks.

As a next step towards methods that are able to automatically generate or optimize recurrent neural networks for a specific task (or class of tasks), it seems worthwhile to combine measures for network properties and task complexity, and devise algorithms that adjust the former based on the latter. 
The approach taken in~\cite{Dambre2012} of using orthogonal functions to measure information processing capacity could be extended to construct suitable dynamical systems for a task when the requirements for a specific task can be measured in a similar way. 

A comparison of how the measures of the information dynamics framework~\cite{Liz07,LPZ12}, the information processing capacity for dynamical systems~\cite{Dambre2012}, measures of criticality~\cite{BN04,PLOW11} or of memory capacity~\cite{Jae01b,GHS08} relate to each other should reveal some interesting insights~\cite[see, e.g.,][]{TR13}, since they all cover some aspects of dynamical systems. Establishing the relation between the information dynamics framework, with recent extension for input-driven systems, and information processing capacity, for example, could help to overcome requirements for i.i.d.\ input in the latter, to better understand dynamical systems with arbitrary input.


\begin{thebibliography}{}

\bibitem[Ashby, 1956]{Ash56}
Ashby, W.~R. (1956).
\newblock {\em An Introduction to Cybernetics}.
\newblock Chapman \& Hall, London.

\bibitem[Baddeley et~al., 1997]{BAB+97}
Baddeley, R., Abbott, L.~F., Booth, M. C.~A., Sengpiel, F., Freeman, T.,
  Wakeman, E.~A., and Roll, E.~T. (1997).
\newblock Responses of neurons in primary and inferior temporal visual cortices
  to natural scenes.
\newblock {\em Proc. R. Soc. Lond. B}, 264:1775--1783.

\bibitem[Bell and Sejnowski, 1995]{BS95}
Bell, A.~J. and Sejnowski, T.~J. (1995).
\newblock An information-maximization approach to blind separation and blind
  deconvolution.
\newblock {\em Neural Computation}, 7(6):1129--1159.

\bibitem[Bengio et~al., 2012]{BBP12}
Bengio, Y., Boulanger-Lewandowski, N., and Pascanu, R. (2012).
\newblock {Advances in Optimizing Recurrent Networks}.
\newblock arXiv preprint 1212.0901, arXiv.org.

\bibitem[Bengio et~al., 1994]{BSF94}
Bengio, Y., Simard, P., and Frasconi, P. (1994).
\newblock {Learning long-term dependencies with gradient descent is difficult}.
\newblock {\em IEEE Transaction on Neural Networks}, 5(2):157--166.

\bibitem[Bertschinger and Natschl{\"a}ger, 2004]{BN04}
Bertschinger, N. and Natschl{\"a}ger, T. (2004).
\newblock Real-time computation at the edge of chaos in recurrent neural
  networks.
\newblock {\em Neural Computation}, 16(7):1413--1436.

\bibitem[Boedecker et~al., 2011]{Boedecker2011}
Boedecker, J., Obst, O., Lizier, J.~T., Mayer, N.~M., and Asada, M. (2011).
\newblock {Information processing in echo state networks at the edge of chaos.}
\newblock {\em Theory in biosciences Theorie in den Biowissenschaften},
  131(3):1--9.

\bibitem[Boedecker et~al., 2009]{Boedecker2009}
Boedecker, J., Obst, O., Mayer, N.~M., and Asada, M. (2009).
\newblock {Initialization and self-organized optimization of recurrent neural
  network connectivity.}
\newblock {\em HFSP journal}, 3(5):340--349.

\bibitem[Crutchfield and Machta, 2011]{CM11}
Crutchfield, J.~P. and Machta, J. (2011).
\newblock Introduction to focus issue on ``{Randomness, Structure, and
  Causality}: Measures of complexity from theory to applications''.
\newblock {\em Chaos}, 21(3):037101.

\bibitem[Dambre et~al., 2012]{Dambre2012}
Dambre, J., Verstraeten, D., Schrauwen, B., and Massar, S. (2012).
\newblock {Information processing capacity of dynamical systems.}
\newblock {\em Scientific reports}, 2:514.

\bibitem[Douglas et~al., 2004]{DMK04}
Douglas, R., Markram, H., and Martin, K. (2004).
\newblock Neocortex.
\newblock In Shepherd, G., editor, {\em Synaptic Organization In the Brain},
  pages 499--558. Oxford University Press.

\bibitem[Doya, 1992]{Doy92}
Doya, K. (1992).
\newblock Bifurcations in the learning of recurrent neural networks.
\newblock In {\em IEEE International Symposium on Circuits and Systems}, pages
  2777--2780. IEEE.

\bibitem[Ganguli et~al., 2008]{GHS08}
Ganguli, S., Huh, D., and Sompolinsky, H. (2008).
\newblock Memory traces in dynamical systems.
\newblock {\em Proceedings of the National Academy of Sciences},
  105(48):18970--18975.

\bibitem[Grassberger, 1986]{Gra86}
Grassberger, P. (1986).
\newblock Toward a quantitative theory of self-generated complexity.
\newblock {\em International Journal of Theoretical Physics}, 25(9):907--938.

\bibitem[Grassberger, 2012]{Gra12}
Grassberger, P. (2012).
\newblock Randomness, information, and complexity.
\newblock Technical Report 1208.3459, arXiv.org.

\bibitem[Hawkins and Blakeslee, 2004]{HB04}
Hawkins, J. and Blakeslee, S. (2004).
\newblock {\em On Intelligence}.
\newblock Times Books.

\bibitem[Hochreiter, 1998]{Hoc98}
Hochreiter, S. (1998).
\newblock {The vanishing gradient problem during learning recurrent neural nets
  and problem solutions}.
\newblock {\em International Journal of Uncertainty, Fuzziness and
  Knowledge-Based Systems}, 6(2):107--116.

\bibitem[Hochreiter and Schmidhuber, 1997]{HS97}
Hochreiter, S. and Schmidhuber, J. (1997).
\newblock Long short-term memory.
\newblock {\em Neural computation}, 9(8):1735--1780.

\bibitem[Jaeger, 2001]{Jae01b}
Jaeger, H. (2001).
\newblock {Short term memory in echo state networks}.
\newblock Technical Report 152, GMD -- German National Research Institute for
  Computer Science.

\bibitem[Jaeger and Haas, 2004]{JH04}
Jaeger, H. and Haas, H. (2004).
\newblock {Harnessing Nonlinearity: Predicting Chaotic Systems and Saving
  Energy in Wireless Communication}.
\newblock {\em Science}, 304(5667):78--80.

\bibitem[Kohonen, 2001]{Koh01}
Kohonen, T. (2001).
\newblock {\em Self-Organizing Maps}.
\newblock Springer, third, extended edition edition.

\bibitem[Kolmogorov, 1965]{Kol65}
Kolmogorov, A.~N. (1965).
\newblock Three approaches to the quantitative definition of information.
\newblock {\em Problemy Peredachi Informatsii}, 1(1):3--11.

\bibitem[Lazar et~al., 2009]{Lazar2009}
Lazar, A., Pipa, G., and Triesch, J. (2009).
\newblock {SORN: a self-organizing recurrent neural network.}
\newblock {\em Frontiers in computational neuroscience}, 3(October):23.

\bibitem[Legenstein and Maass, 2007]{Legenstein2007}
Legenstein, R. and Maass, W. (2007).
\newblock {What makes a dynamical system computationally powerful}.
\newblock In Haykin, S., Principe, J.~C., Sejnowski, T., and McWhirter, J.,
  editors, {\em New Directions in Statistical Signal Processing: From Systems
  to Brains}, pages 127--154. MIT Press.

\bibitem[Linsker, 1987]{Lin87}
Linsker, R. (1987).
\newblock Towards an organizing principle for a layered perceptual network.
\newblock In Anderson, D.~Z., editor, {\em NIPS}, pages 485--494. American
  Institute of Physics.

\bibitem[Lizier et~al., 2007]{Liz07}
Lizier, J.~T., Prokopenko, M., and Zomaya, A.~Y. (2007).
\newblock Detecting non-trivial computation in complex dynamics.
\newblock In {\em Advances in Artificial Life - 9th European Conference on
  Artificial Life (ECAL-2007), Lisbon, Portugal}, volume 4648 of {\em LNAI},
  pages 895--904.

\bibitem[Lizier et~al., 2012]{LPZ12}
Lizier, J.~T., Prokopenko, M., and Zomaya, A.~Y. (2012).
\newblock {Local measures of information storage in complex distributed
  computation}.
\newblock {\em Information Sciences}, 208:39--54.

\bibitem[Luko\v{s}evi\v{c}ius, 2012a]{Luk12}
Luko\v{s}evi\v{c}ius, M. (2012a).
\newblock A practical guide to applying echo state networks.
\newblock In Montavon, G., Orr, G.~B., and M{\"u}ller, K.-R., editors, {\em
  Neural Networks: Tricks of the Trade}, volume 7700 of {\em Lecture Notes in
  Computer Science}. Springer, 2nd edition.

\bibitem[Luko\v{s}evi\v{c}ius, 2012b]{Luk12b}
Luko\v{s}evi\v{c}ius, M. (2012b).
\newblock Self-organized reservoirs and their hierarchies.
\newblock In Villa, A.~E., Duch, W., {\'E}rdi, P., Masulli, F., and Palm, G.,
  editors, {\em Artificial Neural Networks and Machine Learning -- ICANN 2012},
  volume 7552 of {\em Lecture Notes in Computer Science}, pages 587--595.
  Springer Berlin Heidelberg.

\bibitem[Luko\v{s}evi\v{c}ius and Jaeger, 2009]{LJ09}
Luko\v{s}evi\v{c}ius, M. and Jaeger, H. (2009).
\newblock Reservoir computing approaches to recurrent neural network training.
\newblock {\em Computer Science Review}, 3(3):127--149.

\bibitem[Maass et~al., 2007]{MJS07}
Maass, W., Joshi, P., and Sontag, E.~D. (2007).
\newblock Computational aspects of feedback in neural circuits.
\newblock {\em PLOS Computational Biology}, 3(1):e165.

\bibitem[Manjunath et~al., 2012]{Manjunath2012}
Manjunath, G., Tino, P., and Jaeger, H. (2012).
\newblock {Theory of Input Driven Dynamical Systems}.
\newblock In {\em dice.ucl.ac.be}, number April, pages 25--27.

\bibitem[Martens and Sutskever, 2011]{MS11}
Martens, J. and Sutskever, I. (2011).
\newblock Learning recurrent neural networks with hessian-free optimization.
\newblock In {\em Proceedings of the 28th International Conference on Machine
  Learning}, volume~46, page~68. Omnipress Madison, WI.

\bibitem[Martinetz and Schulten, 1991]{MS91}
Martinetz, T. and Schulten, K. (1991).
\newblock A ``neural-gas'' network learns topologies.
\newblock {\em Artificial Neural Networks}, 1:397--402.

\bibitem[Mitchell et~al., 1993]{MHC93}
Mitchell, M., Hraber, P.~T., and Crutchfield, J.~P. (1993).
\newblock Revisiting the edge of chaos: Evolving cellular automata to perform
  computations.
\newblock {\em Complex Systems}, 7:89--130.

\bibitem[Obst et~al., 2010]{Obst2010}
Obst, O., Boedecker, J., and Asada, M. (2010).
\newblock {Improving Recurrent Neural Network Performance Using Transfer
  Entropy}.
\newblock {\em Neural Information Processing Models and Applications},
  6444:193--200.

\bibitem[Obst et~al., 2013]{OBSA13b}
Obst, O., Boedecker, J., Schmidt, B., and Asada, M. (2013).
\newblock On active information storage in input-driven systems.
\newblock preprint 1303.5526v1, arXiv.org.

\bibitem[Ozturk et~al., 2007]{OXP07}
Ozturk, M.~C., Xu, D., and Pr\'{\i}ncipe, J.~C. (2007).
\newblock Analysis and design of echo state networks.
\newblock {\em Neural Computation}, 19(1):111--138.

\bibitem[Prokopenko et~al., 2011]{PLOW11}
Prokopenko, M., Lizier, J.~T., Obst, O., and Wang, X.~R. (2011).
\newblock Relating fisher information to order parameters.
\newblock {\em Physical Review E}, 84(4):041116.

\bibitem[Riedmiller and Braun, 1993]{RB93}
Riedmiller, M. and Braun, H. (1993).
\newblock A direct adaptive method for faster backpropagation learning: the
  rprop algorithm.
\newblock In {\em IEEE International Conference on Neural Networks}, volume~1,
  pages 586--591.

\bibitem[Rissanen, 1978]{Ris78}
Rissanen, J. (1978).
\newblock Modeling by shortest data description.
\newblock {\em Automatica}, 14(5):465--471.

\bibitem[Rolls and Deco, 2010]{RD10}
Rolls, E.~T. and Deco, G. (2010).
\newblock {\em The Noisy Brain - Stochastic Dynamics as a Principle of Brain
  Function}.
\newblock Oxford University Press.

\bibitem[Rumelhart et~al., 1986]{RHW86}
Rumelhart, D., Hinton, G., and Williams, R. (1986).
\newblock Learning representations by back-propagating errors.
\newblock {\em Nature}, 323(6088):533--536.

\bibitem[Schmidhuber et~al., 2007]{SWG+07}
Schmidhuber, J., Wierstra, D., Gagliolo, M., and Gomez, F. (2007).
\newblock Training recurrent networks by evolino.
\newblock {\em Neural computation}, 19(3):757--779.

\bibitem[Schrauwen et~al., 2008]{SWV+08}
Schrauwen, B., Wardermann, M., Verstraeten, D., Steil, J.~J., and Stroobandt,
  D. (2008).
\newblock Improving reservoirs using intrinsic plasticity.
\newblock {\em Neurocomputing}, 71(7-9):1159--1171.

\bibitem[Schreiber, 2000]{Sch00}
Schreiber, T. (2000).
\newblock Measuring information transfer.
\newblock {\em Physical Review Letters}, 85(2):461--464.

\bibitem[Seung, 2012]{Seu12}
Seung, H.~S. (2012).
\newblock {\em Connectome: How the Brain's Wiring Makes Us Who We Are}.
\newblock New York: Houghton Mifflin Harcout.

\bibitem[Sporns, 2011]{Spo11}
Sporns, O. (2011).
\newblock {\em Networks Of the Brain}.
\newblock The MIT Press.

\bibitem[Sussillo and Barak, 2013]{Sussillo2013}
Sussillo, D. and Barak, O. (2013).
\newblock {Opening the black box: low-dimensional dynamics in high-dimensional
  recurrent neural networks.}
\newblock {\em Neural computation}, 25(3):626--49.

\bibitem[Tino and Rodan, 2013]{TR13}
Tino, P. and Rodan, A. (2013).
\newblock {Short term memory in input-driven linear dynamical systems}.
\newblock {\em Neurocomputing}, pages~--.

\bibitem[Triesch, 2005]{Tri05}
Triesch, J. (2005).
\newblock A gradient rule for the plasticity of a neuron's intrinsic
  excitability.
\newblock In Duch, W., Kacprzyk, J., Oja, E., and Zadrozny, S., editors, {\em
  Proceedings of the International Conference on Artificial Neural Networks
  (ICANN 2005)}, Lecture Notes in Computer Science, pages 65--70. Springer.

\bibitem[Voegtlin, 2002]{Voe02}
Voegtlin, T. (2002).
\newblock Recursive self-organizing maps.
\newblock {\em Neural Networks}, 15(8-9):979--991.

\bibitem[Werbos, 1990]{Wer90}
Werbos, P.~J. (1990).
\newblock {Backpropagation through time: what it does and how to do it}.
\newblock {\em Proceedings of the IEEE}, 78(10):1550--1560.

\bibitem[Williams and Beer, 2010]{Williams2010}
Williams, P.~L. and Beer, R.~D. (2010).
\newblock {Information dynamics of evolved agents}.
\newblock In {\em From Animals to Animats 11}, pages 38--49.

\bibitem[Williams and Zipser, 1989]{WZ89}
Williams, R.~J. and Zipser, D. (1989).
\newblock A learning algorithm for continually running fully recurrent neural
  networks.
\newblock {\em Neural Computation}, 1(2):270--280.

\end{thebibliography}
\end{document}